\documentclass[letterpaper, 10 pt, conference]{ieeeconf}  % Comment this line out if you need a4paper

\IEEEoverridecommandlockouts                              % This command is only needed if 
\usepackage{math}
\usepackage{amsmath}
\usepackage{colortbl}
\usepackage[table]{xcolor}
\usepackage{booktabs}
\usepackage[T1]{fontenc}
\usepackage[utf8]{inputenc} % harmless with TL2025; ensures Unicode accents work

\let\labelindent\relax
\usepackage[font=footnotesize, labelfont=footnotesize]{caption}

\usepackage{enumitem}

\newcommand{\best}[1]{\cellcolor{red!30}#1}
\newcommand{\second}[1]{\cellcolor{orange!30}#1}

\title{\LARGE \bf
Splatblox: Traversability-Aware Gaussian Splatting for Outdoor Robot Navigation
}

\author{Samarth Chopra, Jing Liang, Gershom Seneviratne, Yonghan Lee,
Jaehoon Choi, \\ Jianyu An, Stephen Cheng, Dinesh Manocha
\thanks{
All authors are with the University of Maryland, College Park. Corresponding author: {\tt\small sachopra@umd.edu}.
}}

\begin{document}

\maketitle
\thispagestyle{empty}
\pagestyle{empty}
% \linespread{0.98}

%%%%%%%%%%%%%%%%%%%%%%%%%%%%%%%%%%%%%%%%%%%%%%%%%%%%%%%%%%%%%%%%%%%%%%%%%%%%%%%%

% \input{ieeeconf/content_v1}

\begin{abstract}
We present \textit{Splatblox}, a real-time system for autonomous navigation in outdoor environments with dense vegetation, irregular obstacles, and complex terrain. Our method fuses segmented RGB images and LiDAR point clouds using Gaussian Splatting to construct a traversability-aware Euclidean Signed Distance Field (ESDF) that jointly encodes geometry and semantics. Updated online, this field enables semantic reasoning to distinguish traversable vegetation (e.g., tall grass) from rigid obstacles (e.g., trees), while LiDAR ensures 360° geometric coverage for extended planning horizons. We validate \textit{Splatblox} on a quadruped robot and demonstrate its applicability to a wheeled platform. In field trials across vegetation-rich scenarios, it outperforms state-of-the-art navigation methods with over 50\% higher success rate, 40\% fewer freezing incidents, ~5\% shorter paths, and up to 13\% faster time to goal, while supporting long-range missions up to 100\,m. 
Experiment videos and more details can be found on our project page: {\url{https://splatblox.github.io}}

\end{abstract}

%%%%%%%%%%%%%%%%%%%%%%%%%%%%%%%%%%%%%%%%%%%%%%%%%%%%%%%%%%%%%%%%%%%%%%%%%%%%%%%%
\section{INTRODUCTION}
% Autonomous robot navigation in outdoor environments has broad applications including precision agriculture \cite{naik2016precision}, forest exploration \cite{karma2015use}, and search and rescue \cite{borges2022survey}. Such environments present complex challenges due to obstacles of varying shapes, sizes, densities, and opacities, making robust perception and navigation difficult~\cite{sathyamoorthy2023vern}. Outdoor navigation also needs techniques for long-range autonomy~\cite{liang2025mosu}, where robots must plan and execute trajectories that extend well beyond the immediate sensing horizon. 
% A fundamental capability required for robots to operate in these settings is traversability estimation~\cite{mattamala2025wild, e2025terrain, vecchio2024terrain, pan2024traverse}, the ability to distinguish between obstacles that can be traversed (e.g., tall grass~\cite{sathyamoorthy2023vern,ordonez2018characterization}) and those that cannot (e.g., trees or hard bushes).

Autonomous outdoor navigation is critical for applications such as agriculture, forestry, and search and rescue~\cite{naik2016precision,karma2015use,borges2022survey}. These environments contain obstacles with varying geometry, density, and opacity, complicating perception and long-range autonomy beyond the immediate sensing horizon~\cite{sathyamoorthy2023vern,liang2025mosu}. A key capability in such settings is traversability estimation~\cite{mattamala2025wild,e2025terrain,vecchio2024terrain,pan2024traverse}, distinguishing deformable terrain (e.g., tall grass~\cite{sathyamoorthy2023vern,ordonez2018characterization}) from rigid obstacles (such as trees or dense bushes).

Recent work applies deep learning to traversability estimation via supervised, self-supervised, and offline reinforcement learning approaches~\cite{guan2022ga,frey2024roadrunner,wellhausen2019should,mattamala2025wild,weerakoon2023vapor}. While effective in structured settings, these methods require large environment-specific datasets and often rely on robot-dependent supervision tied to particular sensors or dynamics~\cite{elnoor2024amco,villemure2024terrain,seneviratne2024cross}, limiting cross-platform transfer. Offline RL reduces annotation effort but still depends on extensive in-domain trajectory data for generalization~\cite{weerakoon2023vapor}. Moreover, large neural models can be difficult to deploy onboard due to limited GPU memory and inference rates on edge hardware~\cite{weerakoon2024behav}.

\begin{figure}[t!]
    \centering
\includegraphics[width=0.46\textwidth]{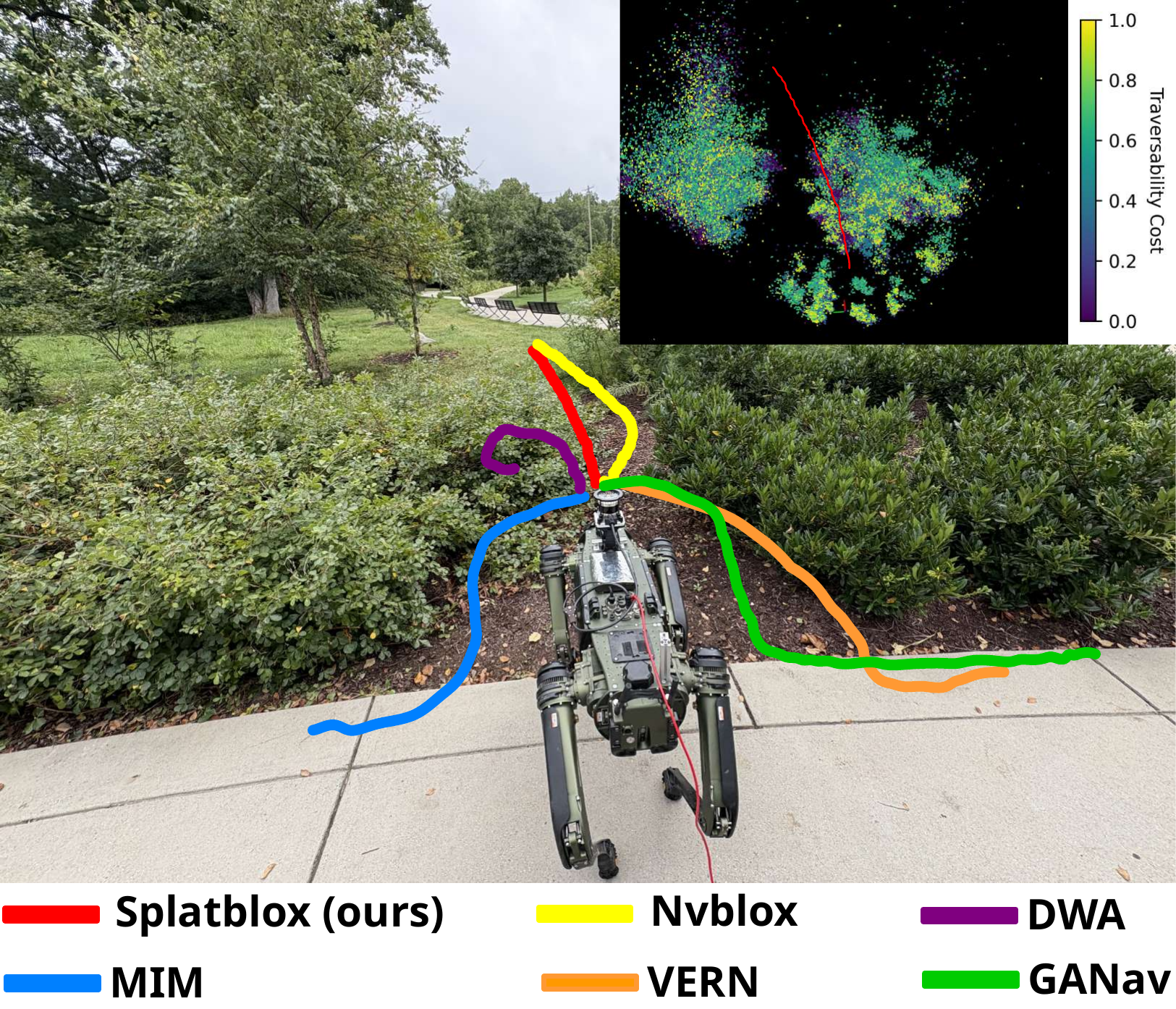}
\caption{
Navigation trajectory of our method \textit{Splatblox} (red) compared to baselines (Nvblox \cite{millane2024nvblox}, DWA \cite{fox2002dynamic}, MIM \cite{sathyamoorthy2024mim}, VERN \cite{sathyamoorthy2023vern}, GA-Nav \cite{guan2022ga}). Our method achieves the lowest normalized trajectory length (NTL) and highest success rate by about 3\% and 60\%, respectively, compared to the second-best method, when navigating through the narrow corridor between bushes (see Table~\ref{table:table_1} for quantitative results). The top-right inset shows the traversability cost volume produced by our GSplat module, with a colormap indicating traversability cost: yellow denotes higher cost (obstacles) and dark blue denotes lower cost (free space).
}
\vspace{-1.0cm}
\label{fig:overview}
\end{figure}

% Learning-free methods LIKE WHAT, GIVE A FEW EXAMPLES circumvent data collection, but are restricted to range-based obstacle detection~\cite{sathyamoorthy2024mim}. Explicit scene representations, such as occupancy grids \cite{elfes2002using}, triangular meshes \cite{edelsbrunner2003surface}, and point clouds \cite{kim2018slam}, provide a geometric representation of the scene, but lack semantic details EXPLAIN WHAT DOES SEMANTIC DETAILS REFERS TO critical for navigation. Implicit representations like Neural Radiance Fields (NeRFs)~\cite{mildenhall2021nerf} can generate photorealistic reconstructions, but they are slow, memory-intensive, and impractical IMPRACTICAL DUE TO WHAT (SLOW SPEED) for path planning. Recently, 3D Gaussian Splatting (GSplats)~\cite{kerbl20233d} has emerged as an efficient alternative FOR WHAT, and can result higher-fidelity reconstructions with faster training WHAT DOES THE TRAINING STEP FOR GAUSSIAN SPLAT RELATED TO YOUR NAVIGATION times~\cite{chen2025splat}.Current GSplat pipelines   require an offline training phase [CITATIONS], and their performance degrades severely on edge devices, often failing due to excessive memory consumption~\cite{mallick2024taming}. Consequently, GSplats have not been deployed onboard robots for real-time outdoor navigation. WHAT ABOUT THIS REFERENCE: https://arxiv.org/abs/2403.02751

% ALSO CITE THIS SURVEY ON USE OF GAUSSIAN SPLATS FOR ROBOTICS

Learning-free methods ~\cite{sathyamoorthy2023vern, sathyamoorthy2024mim, erni2023mem}, avoid the need for labeled datasets but are generally restricted to range-based obstacle detection, without semantic reasoning about vegetation pliability or surface material. Explicit scene representations, including occupancy grids~\cite{elfes2002using}, triangular meshes~\cite{edelsbrunner2003surface}, and point clouds~\cite{kim2018slam}, provide a geometric model of the environment but lack semantic details.
% meaning they do not distinguish between equally geometric obstacles with very different traversability (e.g., grass versus tree trunks).

Implicit neural representations such as Neural Radiance Fields (NeRFs)~\cite{mildenhall2021nerf} can encode both geometry and appearance, producing photorealistic reconstructions, but they are impractical for path planning due to slow training and inference speed as well as high memory demands. In contrast, 3D Gaussian Splatting (GSplat)~\cite{kerbl20233d} has emerged as an efficient alternative to NeRFs for novel view synthesis, achieving faster convergence and higher-fidelity reconstructions. However, the iterative training step in current GSplat pipelines is still time-consuming, which hinders direct use for navigation systems that require online updates~\cite{meuleman2025fly,mallick2024taming}. Moreover, existing implementations often fail on edge devices due to excessive GPU memory consumption~\cite{mallick2024taming}.  

Recent work has begun exploring GSplat for robotics, including safe navigation via splat-based maps~\cite{chen2025splat}, control barrier function filters~\cite{chen2024safer}, and SLAM extensions integrating RGB-D and LiDAR sensing~\cite{matsuki2024gaussian,keetha2024splatam,wei2024gsfusion,lang2024gaussian}. A recent survey further highlights the growing role of Gaussian Splatting in robotics applications~\cite{zhu20243d}. However, these methods are primarily demonstrated in small-scale or indoor environments~\cite{lei2025gaussnav}, and their scalability to real-time, outdoor navigation remains an open challenge~\cite{lang2024gaussian}.

\textbf{Main Contributions:} We present the first real-time Gaussian Splatting navigation system capable of mapping and planning \textit{on-the-fly} in outdoor environments. 
Unlike prior learning-based methods, our approach does not require offline training and generalizes across outdoor scenes with diverse vegetation and terrain, including dense bushes, rocks, trees, mud, mulch, and pavement.
In addition, it runs on resource-constrained edge devices with less than 6\,GB of VRAM at $\approx$2\,Hz reconstruction rates. The novel components of our work include:

\begin{itemize}

 % \item We extend Gaussian Splatting beyond rendering by embedding LiDAR geometry and semantic traversability costs WHAT ARE SEMANTIC TRAVERSABILITY COSTS AND HOW ARE THEY COMPUTED into Gaussian primitives. This results in a fine-grained volumetric field suitable for mapping and navigation in unstructured, vegetation-rich environments.

 \item We extend Gaussian Splatting beyond photorealistic rendering by embedding LiDAR geometry together with semantic traversability costs, derived from a promptable segmentation model that classifies the terrain into traversable or non-traversable categories. This yields a fine-grained volumetric field that encodes both geometry and semantic cues, enabling navigation in vegetation-rich and cluttered environments.
 
 % We demonstrate the first system that integrates RGB and LiDAR sensing into real-time Gaussian Splatting for traversability-aware mapping and navigation in unstructured, vegetation-rich environments.
 
 \item We design a lightweight pipeline optimized for edge GPUs used on mobile robots, which has low memory overhead for real-time performance suitable for field deployment.
 % \item We fuse dense GSplat-based traversability maps with LiDAR-based Euclidean Signed Distance Functions (ESDFs) HOW ARE ESDF COMPUTED, IS THAT A NOVEL CONTRIBUTION, our approach can be used  long-horizon planning that extends beyond the robot’s immediate field of view.  We have used our autononous navigation approach to compute safe trajectories of length up to ?? meters in outdoor environments.
 \item We fuse the GSplat-derived traversability map with LiDAR-based ESDFs, computed using fast GPU-accelerated methods \cite{millane2024nvblox}, to combine semantically rich frontal views with full 360° geometric coverage. This representation supports long-horizon planning up to 100 meters, significantly extending the robot’s local planning horizon.
\end{itemize} 

In our experiments, \textit{Splatblox} consistently outperforms prior methods, including DWA~\cite{fox2002dynamic}, GA-Nav~\cite{guan2022ga}, Nvblox~\cite{millane2024nvblox}, MIM~\cite{sathyamoorthy2024mim}, and VERN~\cite{sathyamoorthy2023vern}, across all evaluation metrics. Compared to the best-performing baseline, our method improves success rate by over 50\%, reduces freezing incidents by 40\%, shortens trajectory length by about 5\%, and decreases time to goal by 13\%. For long-range navigation, \textit{Splatblox} achieves 100\% success rate in scenarios up to 100 meters, while all baselines fail in at least one scenario.

\section{Related Work}
In this section, we give a brief overview of  perception and mapping approaches for outdoor navigation, focusing on three major categories: decoupled perception and planning approaches, end-to-end learning-based methods, and implicit representations such as Gaussian Splatting.

% \subsection{Classical Methods}

\subsection{Perception and Planning (Decoupled Approaches)}

Outdoor navigation has traditionally relied on LiDAR- or stereo-based geometry combined with image-based traversability estimation~\cite{sathyamoorthy2023vern,guan2022ga,sathyamoorthy2024mim,fox2002dynamic}. These approaches typically inflate obstacles or build semantic costmaps fused with LiDAR. While methods such as MIM~\cite{sathyamoorthy2024mim} encode vegetation properties for traversal, most project semantics into 2D ground maps, discarding 3D structure and potentially introducing temporal inconsistencies~\cite{miksik2013efficient}. Hybrid approaches using 2.5D elevation maps~\cite{erni2023mem} or full 3D voxel and octree representations~\cite{hornung2013octomap,oleynikova2017voxblox} preserve geometric structure but can be memory-intensive. Nvblox~\cite{millane2024nvblox} improves efficiency with GPU-accelerated ESDF mapping, yet remains limited to single-modality input.

\subsection{End-to-End Learning-Based Methods}

Deep learning has been widely applied to outdoor navigation through supervised, self-supervised, and offline RL approaches~\cite{guan2022ga,frey2024roadrunner,wellhausen2019should,mattamala2025wild,weerakoon2023vapor,sathyamoorthy2022terrapn}. While supervised methods achieve strong performance, they require costly labeled data, and self-supervised or offline RL methods still depend on large in-domain datasets, limiting scalability and generalization. In practice, many such models are tightly coupled to specific sensor configurations and degrade under sensor shifts~\cite{liang2024mtg,liang2024dtg,liang2025mosu}.

Recent vision-language models (VLMs)~\cite{weerakoon2024behav,sathyamoorthy2024convoi,elnoor2024robot,shah2023lm} enable zero-shot semantic reasoning but impose significant computational overhead, restricting onboard deployment. Similarly, foundation-model-based systems such as ViNT~\cite{shah2023vint} and NoMAD~\cite{sridhar2024nomad} learn topological maps from large trajectory datasets, requiring prior data collection and limiting adaptability in unseen environments. These challenges motivate navigation systems that adapt \textit{on-the-fly} without extensive offline training while remaining computationally efficient for edge deployment.

\subsection{Gaussian Splatting-Based Methods}

Gaussian Splatting (GSplat)~\cite{kerbl20233d} provides an efficient implicit 3D representation with faster convergence than NeRFs, but still requires iterative optimization of Gaussian parameters across frames, limiting real-time deployment in robotics~\cite{mallick2024taming,meuleman2025fly}. Recent works have explored GSplat for mapping, exploration, navigation, and SLAM~\cite{ong2025atlas,tao2024rt,ong2025gaussian,lei2025gaussnav,matsuki2024gaussian,keetha2024splatam,wei2024gsfusion}, as well as safety-aware planning via corridor construction or control barrier filters~\cite{chen2025splat,chen2024safer}. However, these efforts are largely demonstrated in indoor or small-scale settings, and scaling to large outdoor environments remains computationally demanding~\cite{lang2024gaussian,lang2025gaussian}.

% In contrast, we develop a real-time, memory-efficient GSplat framework tailored for outdoor navigation, fusing LiDAR geometry and RGB semantics into a traversability-aware ESDF suitable for onboard planning.

\section{Method}
\begin{figure*}
    \centering
\includegraphics[width=0.88\textwidth]{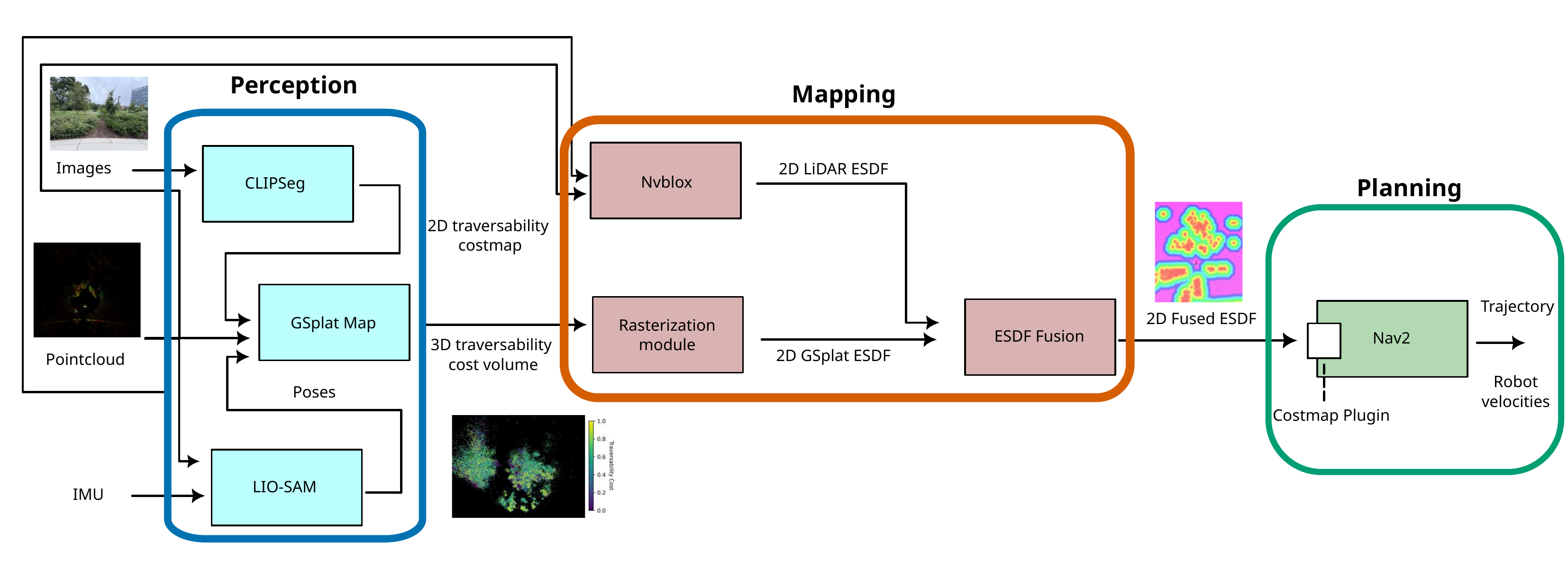}
\vspace{-0.50cm}
\caption{
\textit{Splatblox} architecture: RGB frames are processed by a segmentation module to assign traversability costs to image pixels, which are then associated with LiDAR points projected into the camera frame. These costs and 3D points are used for spawning new Gaussian primitives, forming a traversability-aware volumetric field. The Gaussian Splatting module incrementally updates this field online, maintaining a compact representation on resource-constrained GPUs. From the volumetric field, we derive a GSplat-based Euclidean Signed Distance Field (ESDF) in the robot’s frontal region, which is fused with a LiDAR-based ESDF to ensure 360$^\circ$ geometric coverage. The fused traversability-aware ESDF is provided to the planner as a distance field for collision checking and trajectory generation, enabling the robot to compute safe and efficient paths in vegetation-rich outdoor environments.
}
\vspace{-.8cm}
\label{figure:architecture}
\end{figure*}

%Existing outdoor navigation methods either depend on learning-based models that require extensive training data and fail to generalize, or on geometry-only mapping frameworks that lack the semantic detail needed for safe operation in vegetation-rich environments. To overcome these limitations,
% We propose a real-time Gaussian Splatting system for outdoor navigation that operates \textit{on-the-fly} using RGB and LiDAR sensing. Our approach incrementally reconstructs the scene WHAT DO YOU MEAN BY INCREMENTALLY RECONSTRUCTING A 3D SCENE into a traversability-aware representation and fuses it with LiDAR-based ESDFs to achieve both fine local detail and global coverage EXPLAIN WHAT IS THIS GLOBAL COVERAGE IN SECTION 1. This fused representation is used by the Nav2 planning algorithm~\cite{macenski2020marathon2}, to compute the trajectory.

We propose a real-time Gaussian Splatting system for outdoor navigation that fuses RGB and LiDAR sensing. Our system incrementally reconstructs the scene, inserting and updating Gaussian primitives as new data arrives, to maintain a traversability-aware volumetric map. To extend the planning horizon, this map is fused with a LiDAR-based Euclidean Signed Distance Field (ESDF), combining semantically enriched frontal views with 360° geometric coverage. The fused ESDF is then provided to the robot’s planner for safe trajectory generation in unstructured environments.

\subsection{Traversability Estimation}

We estimate terrain traversability from RGB images using CLIPSeg~\cite{lueddecke22_cvpr}, a promptable segmentation model that enables flexible terrain classification in real time. Given terrain-specific prompts (e.g., ``concrete'', ``sand'', ``rocks''), CLIPSeg produces segmentation masks that are mapped to traversability costs. Following the scheme proposed in~\cite{elnoor2024amco}, we categorize terrain into four classes with increasing cost values in $[0,1]$: (i) \textit{stable} (e.g., concrete), (ii) \textit{granular} (e.g., sand), (iii) \textit{poor foothold/rocky}, and (iv) \textit{high resistance/vegetation}. After assigning the costs, the outputs of the segmentation model are converted into 2D traversability costmaps.

\subsection{Gaussian Splatting as a Traversability-Aware Scene Representation}

% We adopt Gaussian primitives as our scene representation because they are continuous, differentiable, and can be efficiently projected into 2D for fast $\alpha$-blending (WHAT IS ALPHA-BLENDING). This makes them well suited for navigation, by preserving the fine-grained geometric and semantic details required for traversability estimation.
% GSplats represent a scene as an unordered set of 3D Gaussian primitives, which provide a continuous and differentiable approximation of the environment. 
We represent the scene with 3D Gaussian primitives, a continuous and differentiable approximation that preserves fine geometric and semantic detail. Gaussians can be projected into the image plane for fast $\alpha$-blending, that is, weighted compositing of overlapping primitives, which lets us fuse RGB semantic cues with LiDAR geometry.
Each primitive is parameterized by a center $\mu_i \in \mathbb{R}^3$ and covariance $\Sigma_i \in \mathbb{R}^{3 \times 3}$, factorized into scaling $S_i \in \mathbb{R}^3$ and rotation $R_i \in \mathbb{R}^{3 \times 3}$:  

\vspace{-.6cm}
\begin{equation}
    \Sigma_i = R_i S_i S_i^\top R_i^\top.
\end{equation}
\vspace{-.5cm}

This representation is compact and easy to update online, making it well suited for traversability-aware mapping in outdoor environments. In addition, each Gaussian stores an opacity logit $o_i \in \mathbb{R}$ and a color $c_i$ represented by spherical harmonics (SH) coefficients.  While standard GSplat pipelines optimize for photorealistic rendering, in our case projection into the image plane is used to render traversability costmaps, where pixel intensities encode semantic costs rather than color.
Rendering is performed by projecting Gaussians into the image plane, yielding a 2D Gaussian with covariance

\vspace{-.3cm}
\begin{equation}
    \Sigma'_i = J_i W \Sigma_i W^\top J_i^\top,
\vspace{-.2cm}
\end{equation}

where $W \in \mathbb{R}^{3 \times 3}$ is the world-to-camera transform and $J_i \in \mathbb{R}^{2 \times 3}$ the affine projection. Images are reconstructed via $\alpha$-blending of $K$ ordered Gaussians:  

\vspace{-.5cm}
\begin{equation}
    C(x) = \sum_{i=1}^{K} c_i \alpha_i \prod_{j=1}^{i-1}(1-\alpha_j), 
    \quad \alpha_i = o_i \, g^{2D}_i(x).
\end{equation}
\vspace{-.3cm}

To ensure these rendered traversability costmaps align with ground-truth costmaps, we optimize the Gaussian parameters with a combination of pixel-wise and structural losses. Given a rendered map $I$ and ground-truth $I^{gt}$, the objective is  

\vspace{-.3cm}
\begin{equation}
\mathcal{L} = \| I - I^{gt} \|_1 + \big(1 - \text{SSIM}(I, I^{gt})\big).
\vspace{-.2cm}
\end{equation}

Here, the L1 term enforces pixel-level consistency between the predicted and ground-truth traversability costs, while the SSIM (Structural Similarity) term emphasizes structural alignment, preserving boundaries between traversable vegetation and rigid obstacles. 
We do not apply additional depth supervision, as depth is directly derived from metric LiDAR measurements, making an explicit depth loss unnecessary.
% This ensures the learned representation is accurate for navigation, without requiring photorealistic rendering.

% To embed WHAT DO YOU MEAN BY EMBER traversability, 
To incorporate traversability, LiDAR points $\mathbf{p}_i^{\text{lidar}} \in \mathbb{R}^3$ are projected into the camera frame using extrinsic calibration $T_{\text{lidar}\rightarrow\text{cam}} \in SE(3)$:

\vspace{-.4cm}
\begin{equation}
    \mathbf{p}_i^{\text{cam}} = T_{\text{lidar}\rightarrow\text{cam}} \, \mathbf{p}_i^{\text{lidar}}.
\vspace{-.2cm}
\end{equation}

Each point is assigned a traversability cost $c_i \in [0,1]$ from the 2D costmap and incorporated as an anisotropic Gaussian:  
\begin{equation}
    \mathcal{G}_i(\mathbf{x}) = \alpha_i \exp \!\left( -\tfrac{1}{2} (\mathbf{x} - \mathbf{p}_i^{\text{cam}})^\top \Sigma_i^{-1} (\mathbf{x} - \mathbf{p}_i^{\text{cam}}) \right),
\end{equation}
with $\alpha_i$ weighted by $c_i$. The aggregated traversability field is simply $C(\mathbf{x}) = \sum_i \mathcal{G}_i(\mathbf{x})$.

From this volumetric field $C(\mathbf{x})$, we derive an Euclidean Signed Distance Field (ESDF), 
% \begin{equation}
%     d(\mathbf{x}) = \operatorname{sgn}(\mathbf{x}) \min_{\mathbf{y} \in \partial \mathcal{O}} \| \mathbf{x} - \mathbf{y} \|_2,
% \end{equation}
% where $\partial \mathcal{O} = \{ \mathbf{x} \mid C(\mathbf{x}) \geq \tau \}$ defines obstacle boundaries for a threshold $\tau$. 
% The resulting ESDF encodes both geometry and traversability, and is directly used  by the motion planners for safe navigation.  
% The resulting ESDF provides a 
which provides a smooth distance-to-obstacle measure, and is widely used in motion planning for collision checking and trajectory optimization.

% To ensure real-time computation OF WHAT on constrained GPUs, we restrict the number of active Gaussians to $100$k (DO YOU LATER SHOW THE RUNTIME PERFORMANCE AS A FUNCTION OF THE NUMBER OF GAUSSIANS), prioritizing those closest to the camera’s field of view. Remaining Gaussians are either discarded or offloaded to CPU memory aa anchor points, depending on whether offline reconstruction is desired. This design balances reconstruction fidelity with memory efficiency DO YOU PERFORM SOME ANALYSIS LATER IN THE PAPER BETWEEN THE QUALITY OF RECONSRUCTION AND THE RUNTIME, enabling \textit{on-the-fly} deployment on robot-mounted edge devices.

To ensure real-time traversability mapping and ESDF updates on constrained GPUs, we cap the number of active Gaussians at 100k, prioritizing those within or near the camera’s field of view. Gaussians outside this region are discarded. This budget balances reconstruction fidelity with memory efficiency, allowing deployment on robot-mounted edge devices.

\subsection{Euclidean Signed Distance Field (ESDF)}

\begin{figure}[t!]
    \centering
\includegraphics[width=0.46\textwidth]{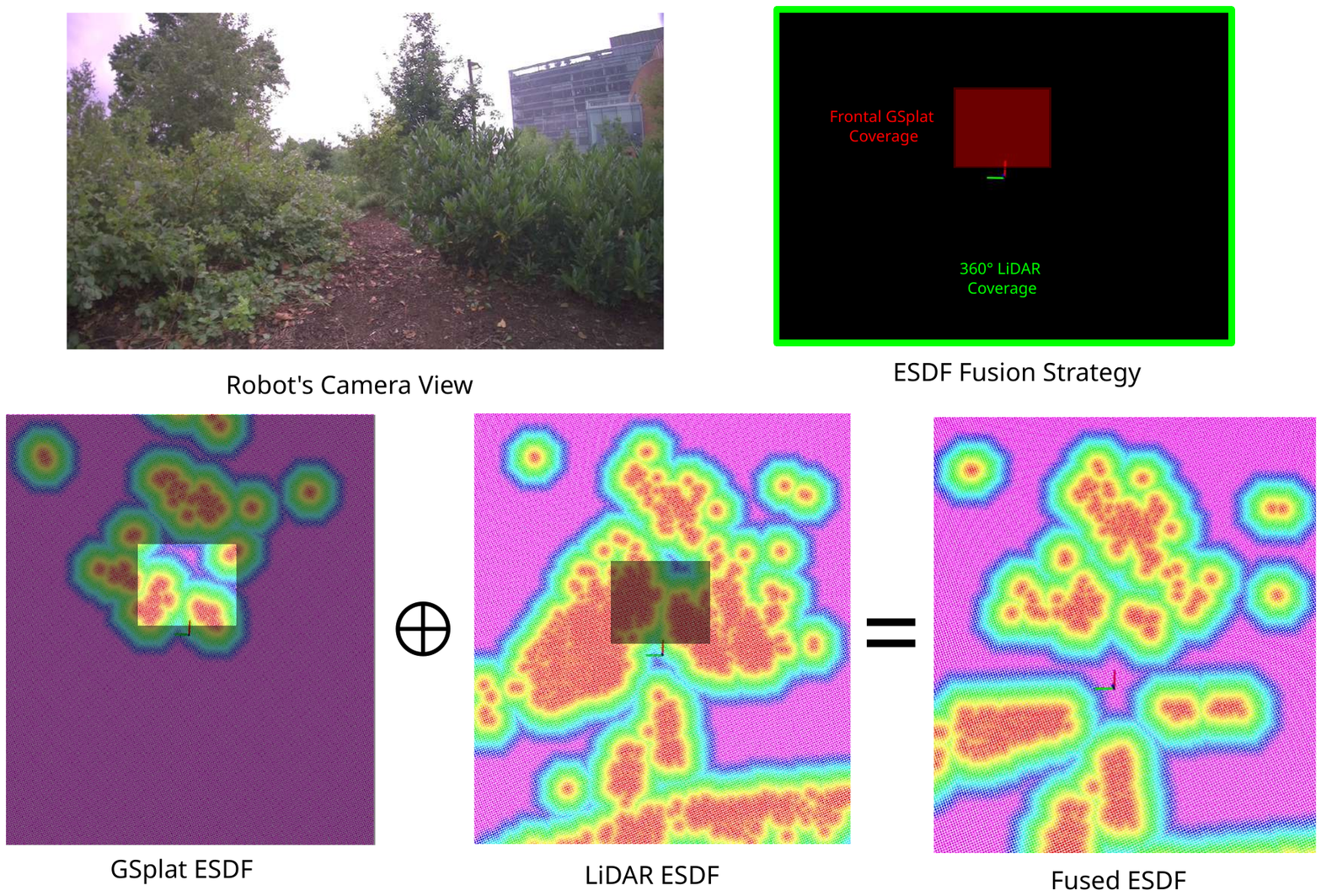}
\vspace{-0.20cm}
\caption{
ESDF fusion strategy. 
% GSplat ESDF provides fine-grained traversability detail in the frontal region (red), while LiDAR ESDF supplies 360$^\circ$ geometric coverage (green). The shaded mask indicates regions of the ESDF that are not used for planning. The fused ESDF combines both ESDFs, enabling the planner to generate collision-free, traversability-aware trajectories. Warmer colors denote regions near obstacles (low distance), cooler colors represent free space (larger distance).
\underline{Top:} Robot’s camera view (left) and fusion strategy (right), where the frontal region is covered by GSplat ESDF (red) and 360° coverage is provided by LiDAR ESDF (green). 
\underline{Bottom:} GSplat ESDF (left) encodes fine-grained traversability detail, LiDAR ESDF (middle) provides global geometric consistency, and their fusion (right) combines both to produce collision-free, traversability-aware trajectories. Warmer colors denote regions near obstacles (low distance), cooler colors represent free space (larger distance).
}
\vspace{-.8cm}
\label{fig:esdf_viz}
\end{figure}

Euclidean Signed Distance Fields (ESDFs) provide a continuous measure of obstacle proximity widely used in mapping and planning. At each point $\mathbf{x} \in \mathbb{R}^n$, the ESDF encodes the signed Euclidean distance to the nearest obstacle boundary $\partial \mathcal{O}$:

\vspace{-.5cm}
\begin{equation}
d(\mathbf{x}) =
\begin{cases}
\;\;\;\min\limits_{\mathbf{y} \in \partial \mathcal{O}} \|\mathbf{x} - \mathbf{y}\|_2, & \mathbf{x} \in \mathbb{R}^n \setminus \mathcal{O}, \\[6pt]
- \min\limits_{\mathbf{y} \in \partial \mathcal{O}} \|\mathbf{x} - \mathbf{y}\|_2, & \mathbf{x} \in \mathcal{O},
\end{cases}
\end{equation}
\vspace{-.4cm}

where positive values indicate free space and negative values indicate points inside obstacles. The gradient $\nabla d(\mathbf{x})$ naturally provides a repulsive vector field, making ESDFs useful both for collision checking and for gradient-based trajectory optimization~\cite{zucker2009chomp,schulman2013trajopt,oleynikova2017voxblox}.  

After the GSplat module reconstructs the scene, we transform the traversability cost volume into a 3D distance field. Specifically, since traversability costs $c \in [0,1]$, we compute the distance field as $(1-c)\,d_{\max}$, where $d_{\max}$ denotes the maximum permissible distance (e.g., $100\,\text{m}$). This scaling ensures that fully traversable regions correspond to large positive distances, while the obstacles yield near-zero values. The resulting volumetric traversability field is then rasterized onto the ground plane to produce a 2D ESDF of the robot's environment, which is used by the planner for collision checking and trajectory generation. 
% To improve safety margins during navigation, we apply a soft inflation kernel to obstacle boundaries, effectively widening their influence region and reducing the likelihood of near-collision maneuvers. 
To improve safety margins during navigation, we apply an inflation kernel to obstacle boundaries with a fixed inflation radius of 0.8 m, effectively widening their influence region and reducing the likelihood of near-collision maneuvers. The inflation radius is kept identical in the Nvblox baseline to ensure fair comparison.

% To provide global coverage WHAT DOES THAT CORRESPOND IN TERMS OF THE REPRESENTATION, we fuse the GSplat-derived ESDF with a LiDAR-based ESDF computed using NVBlox~\cite{millane2024nvblox}. Specifically, we prioritize the GSplat ESDF in a predefined \textit{front grid region} $\mathcal{F} \subset \mathbb{R}^2$ in front of the robot, while using the LiDAR ESDF elsewhere:
% \begin{equation}
% d_{\text{fused}}(\mathbf{x}) =
% \begin{cases}
% d_{\text{GSplat}}(\mathbf{x}), & \mathbf{x} \in \mathcal{F}, \\[6pt]
% d_{\text{LiDAR}}(\mathbf{x}), & \mathbf{x} \notin \mathcal{F}.
% \end{cases}
% \end{equation}
% Our design (DESIGN OF WHAT) leverages camera-based semantic reasoning in the forward region, while ensuring $360^{\circ}$ geometric consistency from LiDAR data (see Figure ???? ).

% The fused traversability-aware ESDF is  provided to the Nav2 navigation framework~\cite{macenski2020marathon2} as a distance field. Within Nav2, this representation is internally converted into a continuous costmap that supports efficient collision checking and trajectory generation. The resulting pipeline enables the robot to plan collision-free, traversability-aware paths that remain robust across both semantically rich frontal views and purely geometric side and rear views.

To extend the planning horizon beyond the camera’s field of view, we fuse the GSplat-derived ESDF with a LiDAR-based ESDF computed using Nvblox~\cite{millane2024nvblox}. Specifically, we prioritize the GSplat ESDF in a predefined front grid region $\mathcal{F} \subset \mathbb{R}^2$ in front of the robot, while using the LiDAR ESDF elsewhere:

\vspace{-.5cm}
\begin{equation}
d_{\text{fused}}(x) = 
\begin{cases} 
d_{\text{GSplat}}(x), & x \in \mathcal{F}, \\ 
d_{\text{LiDAR}}(x), & x \notin \mathcal{F}.
\end{cases}
\end{equation}
\vspace{-.4cm}

This design leverages camera-based semantic reasoning in the frontal view, where traversability cues are most informative, while maintaining 360$^\circ$ geometric consistency from LiDAR (see Fig. \ref{fig:esdf_viz}).

The fused traversability-aware ESDF is then supplied to the planner as a distance field. After an initial map bootstrap phase, mapping and planning proceed concurrently: as new sensor observations are integrated, the ESDF and derived costmap are incrementally updated while the planner simultaneously performs trajectory optimization. Internally, it is converted into a continuous costmap that supports efficient collision checking and trajectory generation. This enables the robot to plan collision-free, traversability-aware paths that remain robust across both semantically rich forward views and purely geometric side and rear views.

\section{Results and Analysis}
\begin{figure*}
    \centering
\includegraphics[width=0.95\textwidth]{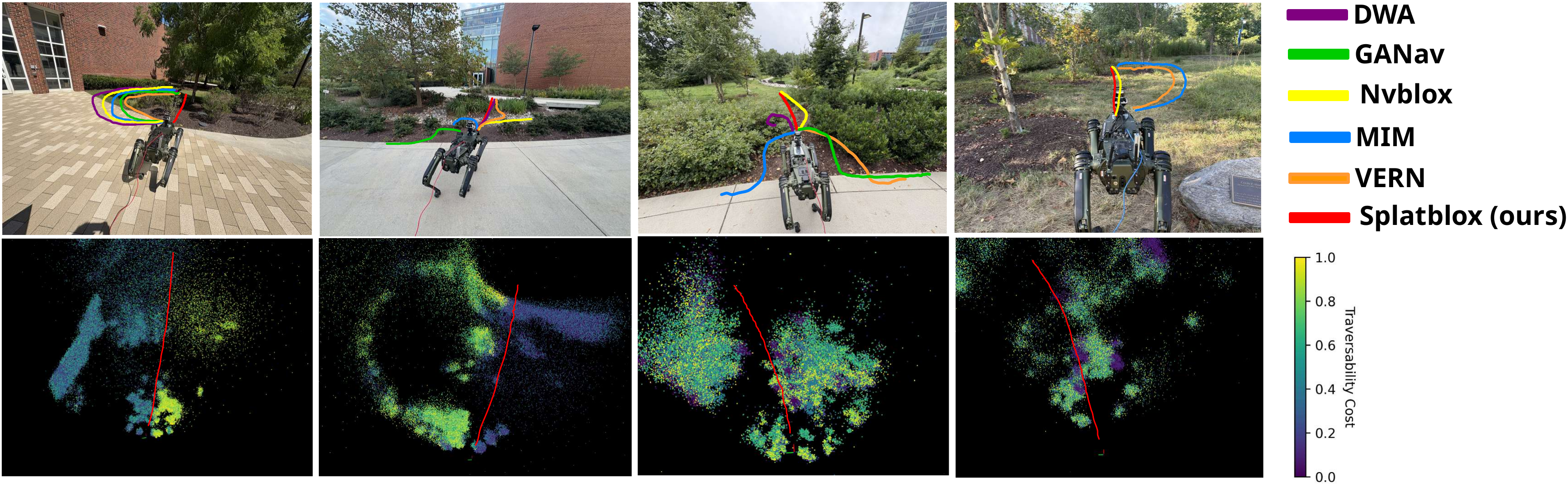}
\caption{
% Navigation trajectories of Splatblox (Ours) and the other methods in the four outdoor scenarios DO YOU KNOW THE GSPLATS FOR THESE SCENES TO SHOW THAT YOUR AGLROTIMS WORKS.
\underline{Top}: Navigation trajectories of our method \textit{Splatblox} (red) compared with baselines (Nvblox \cite{millane2024nvblox}, DWA \cite{fox2002dynamic}, MIM \cite{sathyamoorthy2024mim}, VERN \cite{sathyamoorthy2023vern}, GA-Nav \cite{guan2022ga}) across four outdoor scenarios with diverse terrains, including paved areas, vegetation, and uneven ground. In each case, \textit{Splatblox} achieves successful traversal with the shortest normalized trajectory length (NTL) through narrow passages and cluttered regions, while several baselines either deviate or fail to reach the goal. 
\underline{Bottom}: Traversability cost volume generated by our Gaussian Splatting module (represented as point cloud for visualization) per scene (on top), where yellow indicates high cost (obstacles) and dark blue indicates low cost (traversable). \textit{Splatblox} (red) trajectories are overlaid. Unlike LiDAR-only baselines that treat all objects as obstacles, \textit{Splatblox} leverages semantics to navigate safely through low-cost regions (e.g., grass, bushes), demonstrating the benefit of incorporating semantic cues into the fused ESDF. Overall, \textit{Splatblox} achieves up to 60\% higher success rate (SR), 40\% lower freezing rate (FR), paths up to 7\% shorter (NTL), and time to reach goal (TRG) improvements of up to 13\% compared to baselines.
}
\label{fig:results}
\vspace{-0.70cm}
\end{figure*}

\begin{figure}[t!]
    \centering
\includegraphics[width=0.46\textwidth]{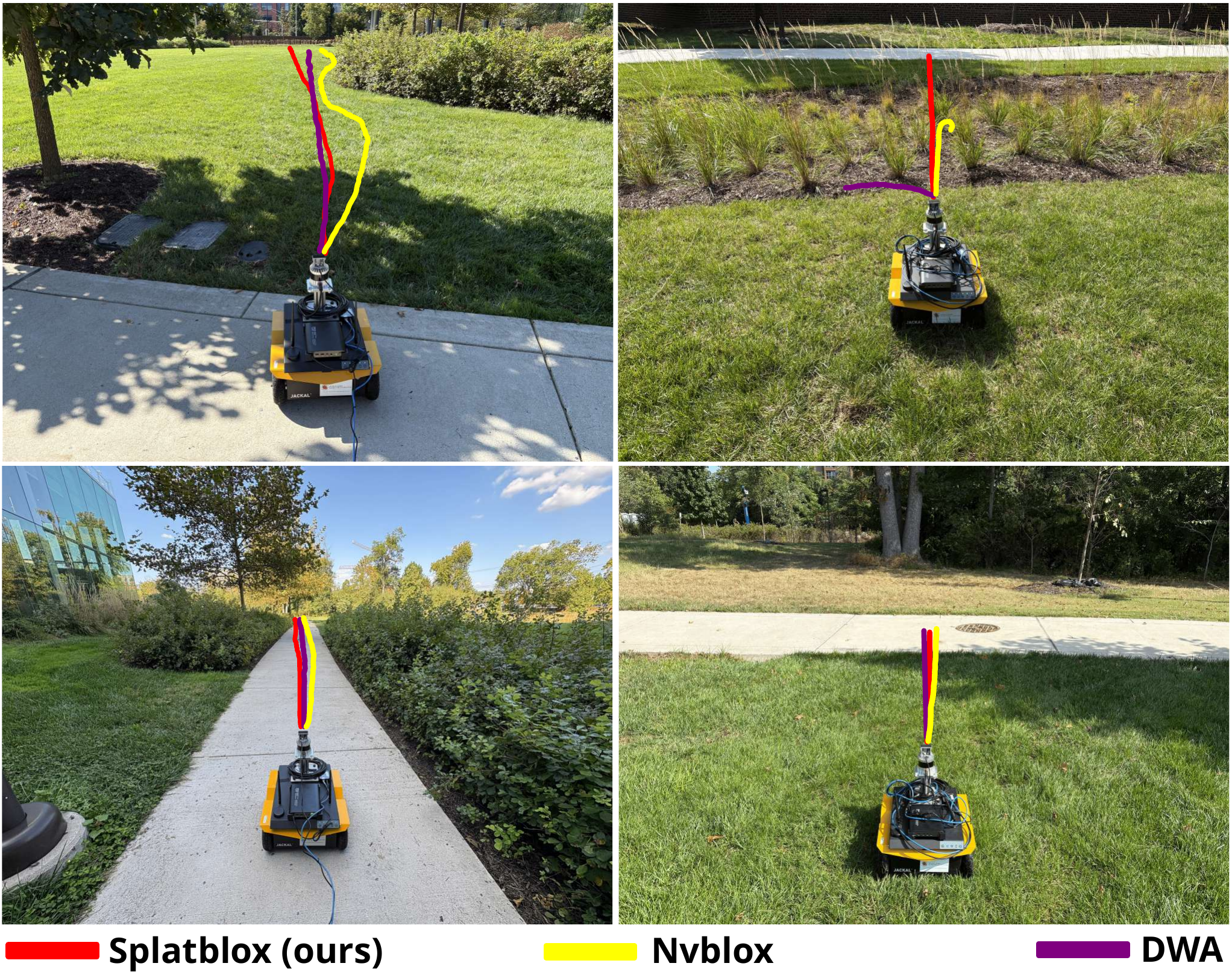}
\caption{
% Longer-range navigation experiments (100 m)
%  \underline{Top}: \textit{Splatblox} (our method) is the only method which is able to succeed in this scenario. The left image depicts the start and the right image depicts the end, where only our method is able to navigate through the vines.
%  \underline{Bottom}: All methods succeed in this scenario, where the scenario starts off on the pavement goes through the grass. Our method places second in normalized trajectory length.
\underline{Top}: \textit{Splatblox} (red) is the only method that successfully completes this scenario. The trajectory begins from the pavement into the grass (left image) and transitions to the vines and shrubs (right image). Our method is the only one able to navigate through the dense vegetation to reach the goal. 
\underline{Bottom}: All methods succeed in a simpler scenario, where the trajectory begins on pavement (left image) and transitions into grass (right image). In this case, \textit{Splatblox} ranks second in normalized trajectory length (NTL) but achieves the fastest time to goal (TRG), improving by 3\% over the best baseline.
}
\label{fig:lrn}
\vspace{-0.90cm}
\end{figure}

\subsection{Implementation Details}
% For outdoor experiments, we deploy our system on the Ghost Vision 60 quadruped robot (Ghost Robotics) as well as Clearpath Jackal UGV, both equipped with a front-facing wide-angle camera, an OS1-32-U LiDAR, and an onboard IMU. Robot poses are estimated using LIO-SAM~\cite{shan2020lio}, which provides both real-time localization and trajectory estimates that are used within the GSplat mapping module. For compute, we use a laptop (connected via ethernet) with an RTX 3070 GPU and an AMD Ryzen 7 5800H CPU.

% Our GSplat implementation builds upon the open-source codebase of Meuleman et al.~\cite{meuleman2025fly}, which we extend to incorporate LiDAR points and camera-derived colors in order to obtain a metric-scale reconstruction of the environment. To compute the LiDAR-based ESDF, we leverage the Nvblox framework~\cite{millane2024nvblox}, and fuse it with our GSplat-derived ESDF obtained via rasterization of the reconstructed scene. The fused representation is then interfaced with the Nav2 navigation framework~\cite{macenski2020marathon2}, where it is converted into a costmap representation suitable for online planning. For motion control, we employ the Model Predictive Path Integral (MPPI) controller \cite{williams2017mppi}, which enables reactive and robust path following in outdoor environments.

Outdoor experiments were conducted on a Ghost Vision 60 quadruped and a Clearpath Jackal UGV, each equipped with a wide-angle camera, OS1-32-U LiDAR, and IMU. Robot poses were estimated using LIO-SAM~\cite{shan2020lio}. We ran the code on a laptop with an RTX 3070 GPU and AMD Ryzen 7 5800H CPU, connected via Ethernet to the robot.

% Our GSplat implementation builds on Meuleman et al.~\cite{meuleman2025fly}, extended to fuse LiDAR points and camera-derived colors for metric-scale reconstruction.
Our GSplat implementation builds upon the optimized real-time 3D Gaussian Splatting framework of Meuleman et al.~\cite{meuleman2025fly}, which uses lightweight keyframe updates with sparse primitive sampling and Sparse Adam optimization. We extend it to fuse LiDAR geometry with camera-derived colors for metric-scale, traversability-aware reconstruction.
LiDAR-based ESDFs were computed with Nvblox~\cite{millane2024nvblox} and fused with GSplat-derived ESDFs via rasterization (using a 5 cm local costmap resolution). The fused field was integrated with Nav2~\cite{macenski2020marathon2} for online planning, and trajectories were executed using the Model Predictive Path Integral (MPPI) controller \cite{williams2017mppi}, which enables reactive and robust path following in outdoor environments. Robot-specific footprints were used for the Ghost (0.85 m x 0.54 m) and Jackal (0.5 m x 0.5 m) to reflect their differing platform dimensions for collision checking. These footprints were applied consistently across methods to ensure fair comparison.

\subsection{Baselines and Metrics}

% For traversability estimation and navigation in diverse terrain, we compare our method to several other navigation methods: DWA (with laserscan) \cite{fox2002dynamic}, GA-Nav \cite{guan2022ga}, Nvblox (with LiDAR) \cite{millane2024nvblox},  MIM \cite{sathyamoorthy2024mim} and VERN \cite{sathyamoorthy2023vern}. We also experiment with using our full method (all Gaussian points) vs just the means of each Gaussian, to verify whether denser representations improve navigation. We run ten experiments per method and per scenario.

% For longer-range navigation (100\,m), we compare our method to DWA (with laserscan) \cite{fox2002dynamic}, and Nvblox (with LiDAR) \cite{millane2024nvblox}. We set 25\,m waypoints between the start and the goal, and run three experiments per method and per scenario.

For traversability estimation and short-range navigation across diverse terrains, we compare \textit{Splatblox} against DWA (laserscan)~\cite{fox2002dynamic}, GA-Nav~\cite{guan2022ga}, Nvblox (LiDAR)~\cite{millane2024nvblox} (using a 5 cm local costmap resolution), MIM~\cite{sathyamoorthy2024mim}, and VERN~\cite{sathyamoorthy2023vern}. These comparisons allow us to disentangle the contribution of each component: Nvblox isolates LiDAR-based geometric mapping without semantic traversability integration, GA-Nav evaluates segmentation-based traversability without volumetric fusion, and DWA serves as a classical reactive baseline. We also evaluate our full method (all Gaussian points) against a variant using only Gaussian means, to test whether denser representations improve navigation. Each method is tested with ten runs per scenario.

For long-range navigation (100\,m), we compare \textit{Splatblox} to DWA~\cite{fox2002dynamic} and Nvblox~\cite{millane2024nvblox}, using 25\,m waypoints between start and goal. Each method is evaluated with three runs per scenario.

We also conduct ablations to evaluate the effect of representation density and grid configuration. Specifically, we vary the proportion of sampled points per Gaussian used for ESDF construction (0\%, 25\%, 50\%, 75\% and 100\%) and observe that denser representations consistently outperform sparser reconstructions. Second, we vary the grid depth (with fixed width of 8m) used for the Gaussian ESDF, allowing us to assess how increased lookahead distance affects navigation performance. Both ablations are performed in scenario~3 of the short-range navigation experiments. Finally, we provide a component-wise breakdown of GPU memory (VRAM) usage and runtime for all system modules.

The metrics we use for evaluation are:

 \begin{itemize}

 \item \textbf{Success Rate (SR)}: The proportion of successful goal-reaching attempts (while avoiding non-pliable vegetation and collisions) over the total number of trials.

 \item \textbf{Freezing Rate (FR)}: The proportion of trials the robot got stuck or started oscillating for more than 10 seconds
 while avoiding obstacles over the total number of attempts. Lower values are better.

 \item \textbf{Normalized Trajectory Length (NTL)}: The ratio between the robot’s trajectory length and the straight-line distance to the goal in all the successful trajectories.

 % \item \textbf{False Positive Rate (FPR)}: The ratio between the number of false positive predictions (i.e., actually untraversable/non-pliable obstacles predicted as traversable) and the total number of actual negative (untraversable) obstacles encountered during a trial. We report the average over all the trials.

 \item \textbf{Time to Reach Goal (TRG)}: The time taken to reach the goal in seconds. Only successful attempts are counted.

 \end{itemize}

\subsection{Testing Scenarios}

Following are the scenarios we test for traversability estimation:

\begin{itemize}

\item \textbf{Scenario 1}: Contains narrow passages between bushes and a tree in a mulch surface.

\item \textbf{Scenario 2}: Shrubs, and trees with narrow openings and a rocky surface.

\item \textbf{Scenario 3}: Dense bushes and shrubs with a narrow opening.

\item \textbf{Scenario 4}: Long grass with bushes and trees.

\end{itemize}

Following are the scenarios we test for longer range missions:

\begin{itemize}

\item \textbf{Scenario 1}: Tree, bushes, large lawn with vines and shrubs.

\item \textbf{Scenario 2}: Pavement, bushes and lawn.

\end{itemize}

\subsection{Analysis and Comparison}

\begin{table}[t]
\centering
\caption{Short-range navigation performance across diverse terrains. Best results are shown in \textbf{red}, and second-best in \textbf{orange}.
}
\label{table:table_1}
% EXPLAIN THE METRICS IN THE TABLE AND THE RUNNING TIME.
\renewcommand{\arraystretch}{1.1} % reduce row height a bit
\setlength{\tabcolsep}{3pt}      % reduce horizontal padding

% Resize the table to fit within column width
\resizebox{\columnwidth}{!}{%
\begin{tabular}{l lcccc}
\toprule
 & Method & SR$\uparrow$ (\%) & NTL & FR$\downarrow$ (\%) & TRG$\downarrow$ (s) \\
\midrule

\multirow{7}{*}{\textbf{Scen. 1}} 
 & DWA (with laserscan) \cite{fox2002dynamic}  & 70 & 1.42 & 30  & 21.0 \\
 & GANav \cite{guan2022ga}                     & 50  & 1.32  & 50  & 19.5 \\
 & Nvblox (with LiDAR) \cite{millane2024nvblox}                        & 50  & 1.35  & \best{0}  & 20.6 \\
 & MIM \cite{sathyamoorthy2024mim}             & \best{100}  & 1.33  & \best{0}  & 16.1 \\
 & VERN \cite{sathyamoorthy2023vern}           & \best{100}  & 1.31  & \best{0}   & 14.0 \\
 & Ours (Only Gaussian means)                  & 80 & \best{1.20} & \second{10}  & \best{12.1} \\
 & Ours (All Gaussian points)                  & \second{90} & \second{1.26}  & \second{10}  & \second{12.2} \\
\midrule

\multirow{7}{*}{\textbf{Scen. 2}} 
 & DWA (with laserscan) \cite{fox2002dynamic}  & \second{90}  & 1.16  & 0  & \second{11.5} \\
 & GANav \cite{guan2022ga}                     & 0  & -  &  100 & - \\
 & Nvblox (with LiDAR) \cite{millane2024nvblox}                        & 0  & -  & 100  & - \\
 & MIM \cite{sathyamoorthy2024mim}             &  0 & -  & 100  & - \\
 & VERN \cite{sathyamoorthy2023vern}           & 20  & 1.65  & \best{0}  & 22.0 \\
 & Ours (Only Gaussian means)                  &  \best{100} & \best{1.04}  & \best{0} & \best{10.8} \\
 & Ours (All Gaussian points)                  & \best{100}  & \second{1.12}  &  \best{0} & \best{10.8} \\
\midrule

\multirow{7}{*}{\textbf{Scen. 3}} 
 & DWA (with laserscan) \cite{fox2002dynamic}  &  0 & -  & 100  & - \\
 & GANav \cite{guan2022ga}                     &  0 & -  & 100  & - \\
 & Nvblox (with LiDAR) \cite{millane2024nvblox}                         &  40 & \second{1.05}  & \second{60}  & 14.8 \\
 & MIM \cite{sathyamoorthy2024mim}             &  0 & -  & 100  & - \\
 & VERN \cite{sathyamoorthy2023vern}           &  0 & -  & 100  & - \\
 & Ours (Only Gaussian means)                  &  \second{90} & \best{1.02}  & \best{0}  & \best{13.8} \\
 & Ours (All Gaussian points)                  &  \best{100} & \best{1.02}  & \best{0}  & \second{14.2} \\
\midrule

\multirow{7}{*}{\textbf{Scen. 4}} 
 & DWA (with laserscan) \cite{fox2002dynamic}  &  0 & -  & 100  & - \\
 & GANav \cite{guan2022ga}                     &  0 & -  & 100  & - \\
 & Nvblox (with LiDAR) \cite{millane2024nvblox}                        & \second{90}  & 1.15  & 10  & 17.3 \\
 & MIM \cite{sathyamoorthy2024mim}             &  30 & 1.65  & 70  & 43.0 \\
 & VERN \cite{sathyamoorthy2023vern}           &  10 & 1.61  & 90  & 34.0 \\
 & Ours (Only Gaussian means)                  &  80 & \best{1.06}  & \second{0}  & \best{16.7} \\
 & Ours (All Gaussian points)                  &  \best{100} & \second{1.07}  & \best{0}  & \second{17.0} \\
\bottomrule
\vspace{-0.7cm}
\end{tabular}
}% end resizebox

\end{table}

\begin{table}[t]
\centering
\caption{Long-range (100\,m) navigation performance across outdoor scenarios. Best results are shown in \textbf{red}, and second-best in \textbf{orange}.}
\label{table:lrn}
\renewcommand{\arraystretch}{1.1} % reduce row height a bit
\setlength{\tabcolsep}{3pt}      % reduce horizontal padding

% Resize the table to fit within column width
\resizebox{\columnwidth}{!}{%
\begin{tabular}{l lcccc}
\toprule
 & Method & SR$\uparrow$ & NTL & FR$\downarrow$ & TRG$\downarrow$ \\
\midrule

\multirow{3}{*}{\textbf{Scen. 1}} 
 & DWA (with laserscan) \cite{fox2002dynamic}  &   0 & -  & 100 & - \\
 & Nvblox (with LiDAR) \cite{millane2024nvblox} & 0  & -  & 100  & - \\
 & Ours (All Gaussian points)                  & \best{100}  & \best{1.05}  & \best{0}  & \best{215.0} \\
\midrule

\multirow{3}{*}{\textbf{Scen. 2}} 
 & DWA (with laserscan) \cite{fox2002dynamic}  &   \best{100} & \best{1.00}  & \best{0}  & 155.0 \\
 & Nvblox (with LiDAR) \cite{millane2024nvblox} & \best{100} & 1.05  & \best{0}  & \second{145.0} \\
 & Ours (All Gaussian points)                  & \best{100}  & \second{1.02}  & \best{0}  & \best{140.0} \\
\bottomrule
% \vspace{-1.0cm}
\vspace{-0.7cm}
\end{tabular}
}% end resizebox

\end{table}

\begin{figure}[t!]
    \centering
\includegraphics[width=0.46\textwidth]{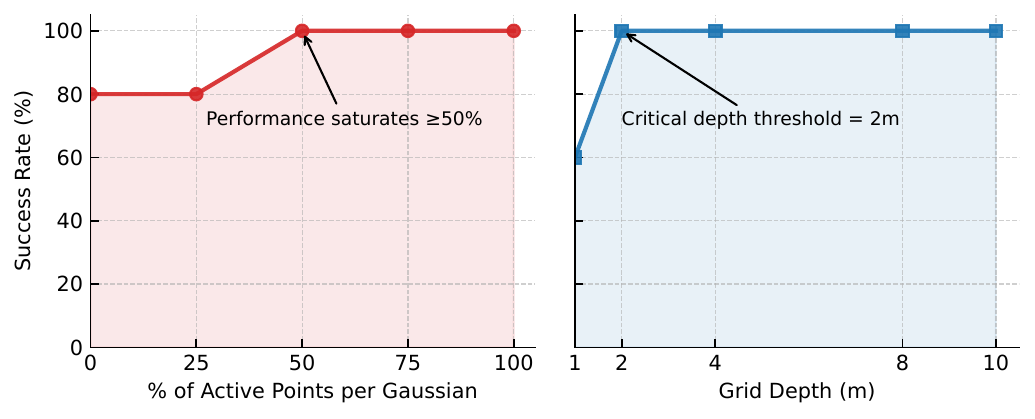}
\vspace{-0.20cm}
\caption{
 \underline{Left}: Success rate as a function of the proportion of sampled points per Gaussian. Increasing GSplat density yields more fine-grained traversability representations, improving navigation performance, with gains saturating beyond 50\%. \underline{Right}: Success rate as a function of Gaussian grid depth, showing a critical threshold at 2\,m. Increasing the lookahead distance improves navigation, with success rate rising by +40\% when grid depth increases from 1 to 2\,m.
}
\vspace{-0.80cm}
\label{fig:ablations}
\end{figure}

\begin{table}[t]
\centering
\caption{Component-wise GPU memory ($\Delta$ VRAM) and runtime breakdown.}
\label{tab:compute_breakdown}
\small
\setlength{\tabcolsep}{4pt}
\renewcommand{\arraystretch}{0.95}
\begin{tabular}{lcc}
\toprule
\textbf{Component} & \textbf{$\Delta$ VRAM (MB)} & \textbf{Runtime (ms)} \\
\midrule

\multicolumn{3}{l}{\textbf{GSplat}} \\
Add Gaussians (100k) & 4000 & 51.82 \\
Publish GS--LiDAR ESDF & 0 & 56.80 \\
Add Keyframe & 0 & 8.11 \\
Gaussians $\rightarrow$ Pointcloud & 0 & 39.02 \\
Keyframe Optimization & 0 & 3.57 \\
\textbf{GSplat Total} & \textbf{4000} & \textbf{159.32} \\
\midrule
\multicolumn{3}{l}{\textbf{Nvblox}} \\
ESDF Update & 609 & 204.08 \\

\multicolumn{3}{l}{\textbf{CLIPSeg}} \\
Semantic Inference & 640 & 479.26 \\

\midrule
\textbf{System Total (Parallel)} & \textbf{5249} & \textbf{479.26} \\
Effective Rate & -- & 2.09 Hz \\
\bottomrule
\end{tabular}
\vspace{-0.7cm}
\end{table}

\subsubsection{Traversability Estimation and Short Range Navigation}

In \textbf{Scenario 1}, our method (both means and all points) is the only one able to navigate through the narrow corridor between bushes, while others detour around them to reach the goal. This highlights our ability to identify traversable regions with fine-grained detail. The full method achieves the second-best SR and FR (–10\% compared to MIM and VERN), but obtains the shortest path (NTL 1.26 vs.\ 1.31) and a +12.9\% faster TRG compared to the second-best method (see Table~\ref{table:table_1}).

In \textbf{Scenario 2}, our method achieves the best overall performance with the highest SR, lowest NTL (1.04 vs.\ 1.16), lowest FR, and a 6.1\% improvement in TRG. As in Scenario 1, our method reliably identifies the traversable corridor and follows it to the goal.

In \textbf{Scenario 3}, our method again prevails across all metrics (SR, NTL, FR, TRG). Both the full points and means variants are able to find the corridor through dense bushes, while the best competing method, Nvblox~\cite{millane2024nvblox}, succeeds only 40\% of the time.

In \textbf{Scenario 4}, our method consistently outperforms baselines, achieving the highest SR, lowest FR, and shortest traversable paths to the goal.

% Across all scenarios, the full method (all Gaussian points) consistently attains higher SR than the Gaussian-means variant, as trajectories generated from means alone sometimes collide with obstacles. The all-points representation is more conservative due to its denser and more precise scene encoding. Consequently, its NTL and TRG are generally equal to or slightly higher than the means variant. In other words, the full method produces safer, more reliable trajectories at the cost of marginally longer paths and runtimes.

Across all scenarios, the full method (all Gaussian points) achieves higher SR than the Gaussian-means variant, as using means alone can lead to collisions. The denser representation yields more conservative planning, resulting in marginally higher NTL and TRG but more reliable trajectories.

% The advantage of Gaussian splatting extends beyond occupancy alone. Sparse LiDAR returns provide discrete surface samples, whereas alpha blending yields a continuous and volumetrically denser scene representation, reducing gaps and discretization artifacts in the ESDF, particularly around thin structures and vegetation. Even compared to Nvblox with a 5 cm local costmap resolution, this improved geometric continuity enables reliable navigation through narrow traversable corridors, with the largest gains observed in Scenarios 2 and 3.

The advantage of Gaussian splatting extends beyond occupancy alone. Sparse LiDAR returns provide discrete surface samples, whereas alpha blending yields a continuous and volumetrically denser scene representation that reduces gaps and discretization artifacts in the ESDF, particularly around thin structures and vegetation. Methods relying on LiDAR-based voxel ESDFs (Nvblox, 5 cm local costmap resolution) or segmentation-based traversability without volumetric fusion (GA-Nav) struggle in complex terrain, especially in Scenarios 2 and 3. We observe that GA-Nav assigns higher costs to dense vegetation and consequently favors paved surfaces, highlighting the advantage of a promptable model such as CLIPSeg for traversability-aware navigation in unstructured environments. Our splat-based representation fuses semantic costs directly in 3D prior to ESDF construction, supporting reliable navigation through narrow traversable corridors where other methods fail.

% In \textbf{Scenario 1}, \textit{Splatblox} (both variants) is the only method to traverse the narrow corridor between bushes, while baselines detour around obstacles. Although the full method achieves slightly lower SR (–10\% vs.\ MIM/VERN), it yields shorter paths (NTL 1.26 vs.\ 1.31) and 12.9\% faster TRG (Table~\ref{table:table_1}).

% In \textbf{Scenarios 2–4}, \textit{Splatblox} consistently outperforms baselines across all metrics, achieving 100\% SR in Scenarios 2 and 4, and improving SR by +60 \% in Scenario 3 (40\% $\to$ 100\%). Qualitatively, our method reliably identifies traversable corridors through vegetation where others fail. Across all cases, the full Gaussian representation is more conservative than using only means, producing safer trajectories at the cost of slightly longer NTL and TRG.

\subsubsection{Longer-range navigation (100\,m)}

In \textbf{Scenario 1}, our method is the only one able to successfully traverse through vines and shrubs, demonstrating robust traversability estimation (see Fig.~\ref{fig:lrn} and Table~\ref{table:lrn}). In contrast, Nvblox (LiDAR)~\cite{millane2024nvblox} and DWA (laserscan)~\cite{fox2002dynamic} treat these regions as solid obstacles and freeze.

In \textbf{Scenario 2}, all methods succeed (100\% SR), as the task involves navigating along pavement and then crossing grass to reach the goal. While our method ranks second in NTL, it outperforms in all other metrics. The higher TRG performance arises because DWA, despite achieving the lowest NTL, progresses slowly and cautiously along the pavement corridor, resulting in longer execution time.

\subsubsection{Ablations}

Increasing the proportion of sampled points per Gaussian used to construct the 3D traversability cost volume improves SR monotonically (see Fig.~\ref{fig:ablations}). Denser Gaussians provide finer scene detail than means alone, enabling more accurate traversability estimation and safer navigation. The full method, which leverages all points, consistently surpasses the means-only variant, though performance saturates beyond 50\% of active points.

Increasing the Gaussian grid depth also improves SR by extending the effective lookahead distance. At a depth of 1\,m, our method experiences two collisions with trees, while at 2\,m the SR improves from 60\% to 100\%. This effect arises because our reconstruction runs at $\approx$2\,Hz; if the robot moves faster than this rate, blind regions appear in the Gaussian grid and cause failures. We identify 2\,m as a critical threshold, beyond which additional depth does not provide further gains.

\subsubsection{Compute breakdown}

Table~\ref{tab:compute_breakdown} provides a component-wise GPU memory and runtime breakdown of the three parallel nodes. The overall system runs at 2.09 Hz, compared to 4.9 Hz for Nvblox under the same hardware conditions. While this is below the update rates of reactive planners such as DWA ($\approx$10 Hz with LiDAR) and GANav ($\approx$20 Hz with camera), which operate near sensor rate, our pipeline incorporates semantic segmentation, Gaussian scene modeling, and ESDF fusion within the planning loop. Performance is primarily bottlenecked by CLIPSeg, which runs as an independent node and evaluates 14 prompts per frame; reducing or generalizing these prompts could significantly improve runtime for dynamic deployments. Using four general-purpose prompts (“concrete”, “grass”, “bush”, and “dirt”) reduces semantic inference time to 309 ms (-35.6\%) and increases the update rate to 3.16 Hz (+51.2\%). Substituting CLIPSeg with a more efficient pretrained model such as GANav could further increase throughput.

\section{Conclusions, Limitations and Future Work}
We presented \textit{Splatblox}, a real-time Gaussian Splatting system for traversability-aware navigation in outdoor environments. Our method fuses RGB-based semantic costs with LiDAR geometry into a unified volumetric representation, which is converted into an ESDF and supplied to the planner. Across diverse terrains and 100\,m long-horizon scenarios, Splatblox improves success rate by up to 60\% over the best competing methods, while also reducing trajectory length by up to 7\%, improving time-to-goal by up to 13\%, and achieving substantially lower freezing rates. Ablation studies further confirm the importance of dense Gaussian representations and sufficient grid depth for fine-grained traversability estimation, and thus safe and reliable navigation. 

Despite these results, several limitations remain.  
% First, our system currently runs at $\approx$2 Hz, which restricts deployment to largely static environments; it cannot yet handle dynamic scenes. 
% Second, our reconstruction relies solely on LiDAR for pointcloud input. While effective for geometry, incorporating complementary, denser depth sources such as RGB-D cameras could improve reconstruction fidelity, photorealism, and ultimately traversability estimation.  
% Third, our approach is not robust under degraded lighting conditions, as the underlying costmaps are derived from vision-based segmentation. Although LiDAR provides local geometric awareness, the traversability pipeline fails when the segmentation model degrades. In future work, we plan to incorporate additional sensing modalities to improve robustness in low-light and other challenging conditions. Overall, our results suggest that Gaussian Splatting can serve as a foundation for reliable and efficient outdoor navigation.
The current system operates at $\approx$2 Hz and is best suited to quasi-static environments. Reconstruction relies primarily on LiDAR pointcloud input and performance degrades under challenging lighting conditions due to vision-based segmentation. Additionally, occlusions and partial observability may lead to incomplete map information, potentially affecting reliable collision avoidance. Future work will focus on improving runtime, incorporating additional sensing modalities, and extending the framework to dynamic environments through motion prediction. Overall, our results demonstrate that Gaussian Splatting provides a promising foundation for efficient, traversability-aware outdoor navigation.

\section*{ACKNOWLEDGMENT}

We thank the research staff at the Maryland Robotics Center (MRC), Ivan Penskiy, for their assistance in providing access to the Jackal robot and ensuring that all hardware and software systems were operational for our experiments.

\bibliographystyle{IEEEtran}
\bibliography{literature}

\end{document}